%% file: arxiv.tex
\documentclass[10pt,twocolumn,letterpaper]{article}

\usepackage[pagenumbers]{iccv} 


\definecolor{iccvblue}{rgb}{0.21,0.49,0.74}
\usepackage[pagebackref,breaklinks,colorlinks,allcolors=iccvblue]{hyperref}

\def\paperID{14440} 
\def\confName{ICCV}
\def\confYear{2025}

\title{Wavelet Policy: Lifting Scheme for Policy Learning in Long-Horizon Tasks}

\author{%
  Hao Huang\thanks{Equal contribution: {hh1811,sy2366}@nyu.edu.}\enspace$^{1,3}$, Shuaihang Yuan$^\ast$$^{1,2,3}$, Geeta Chandra Raju Bethala$^{1,3}$, Congcong Wen$^{1,3}$ \and Anthony Tzes$^{2,3}$, Yi Fang
  \thanks{Corresponding author: yfang@nyu.edu.}\enspace$^{1,2,3}$\\
  \\
  $^{1}$Embodied AI and Robotics (AIR) Lab, NYUAD\\
  $^{2}$NYUAD Center for Artificial Intelligence and Robotics (CAIR)\\
  $^{3}$New York University Abu Dhabi\\
}

\begin{document}
\maketitle

\begin{abstract}
Policy learning focuses on devising strategies for agents in embodied artificial intelligence systems to perform optimal actions based on their perceived states. One of the key challenges in policy learning involves handling complex, long-horizon tasks that require managing extensive sequences of actions and observations with multiple modes. Wavelet analysis offers significant advantages in signal processing, notably in decomposing signals at multiple scales to capture both global trends and fine-grained details. In this work, we introduce a novel wavelet policy learning framework that utilizes wavelet transformations to enhance policy learning. Our approach leverages learnable multi-scale wavelet decomposition to facilitate detailed observation analysis and robust action planning over extended sequences. We detail the design and implementation of our wavelet policy, which incorporates lifting schemes for effective multi-resolution analysis and action generation. This framework is evaluated across multiple complex scenarios, including robotic manipulation, self-driving, and multi-robot collaboration, demonstrating the effectiveness of our method in improving the precision and reliability of the learned policy. Our project is available at \url{https://hhuang-code.github.io/wavelet_policy/}.
\end{abstract}

\section{Introduction}
\label{sec:intro}
Policy learning in embodied artificial intelligence aims to generate optimal actions for an agent to achieve specific goals based on the observed states. This field has evolved from foundational methods such as behavior cloning~\cite{ho2016generative,torabi2018behavioral} and reinforcement learning~\cite{pinto2017robust,kaisermodel} to more complex approaches like hierarchical reinforcement learning~\cite{hafner2022deep,lee2023adaptive} and inverse reinforcement learning~\cite{haldar2023watch,swamy2023inverse}. Recent trends in policy learning emphasize handling complex, multi-stage tasks that require high precision operation and long-term decision-making in environments with multi-modal action distributions and high-dimensional action spaces~\cite{gupta2020relay,shafiullah2022behavior,chi2023diffusion}. Techniques such as implicit policy learning~\cite{florence2022implicit} which generates agent behavior through an implicit energy-based model~\cite{lecun2006tutorial} and hierarchical visual policy learning~\cite{wang2023hierarchical} tailored for long-horizon tasks in cluttered scenes, have illustrated the breadth of policy learning being developed.

\begin{figure}[t]
\centering
\includegraphics[width=.99\linewidth]{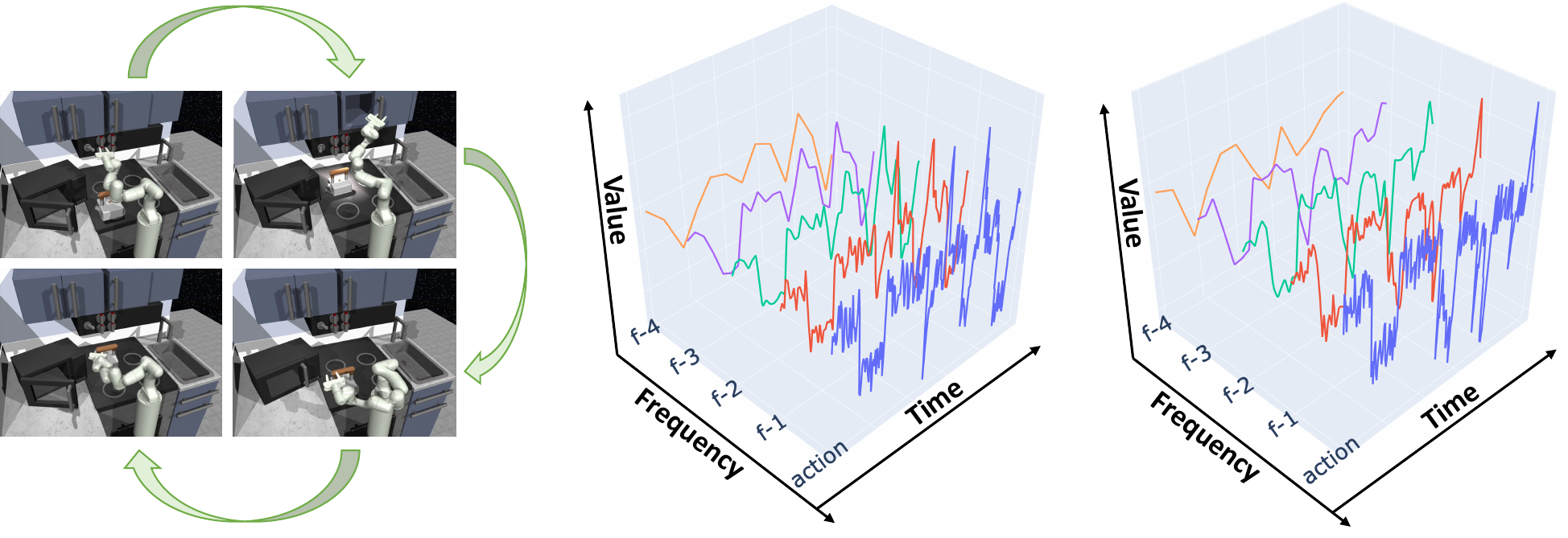}
\caption{A kitchen robot arm (left) conducts a series of actions, and we decompose the action sequences (blue) of the arm's $8^{th}$ (middle) and $9^{th}$ (right) joints into four frequencies ($f\text{-}1$ to $f\text{-}4$, red to orange) using Haar wavelet. Here $f\text{-}1$ represents high-frequency components of the original actions, while $f\text{-}4$ represents the low-frequency of the actions. The action modes (trends) become clear in the low-frequency components without fluctuations (noises).}
\label{fig:teaser}
\end{figure}

Despite the advances mentioned above, policy learning for embodied agents still faces significant challenges, particularly when dealing with complex, long-horizon tasks that require handling long observation and action sequences~\cite{sermanet2021broadly,chensequential}. In long-horizon scenarios, an agent must maintain consistent behavior across multiple steps, effectively managing dependencies over time, which can lead to an accumulation of errors if not properly handled. Furthermore, many real-world tasks exhibit multi-modal action patterns, \ie, there are often multiple valid action sequences to achieve the same goal~\cite{shafiullah2022behavior,leebehavior}. This diversity in valid actions creates more difficulties for the learning process, which must generalize across different potential action plans without overfitting to a specific one. Lastly, certain applications require high precision in operations~\cite{morgan2021vision,drolet2024comparison}, as small inaccuracies in control or planning can lead to significant issues, especially in delicate environments like medical diagnosis or industrial automation. Therefore, ensuring the learned policies to generate reliable, precise actions under complex conditions is one of the most challenging yet important aspects of policy learning.

Wavelet analysis~\cite{daubechies1993ten} is a powerful tool in signal processing by decomposing signals into multiple scales, capturing both global trends and localized details in time and frequency domains. Inspired by such characteristics of wavelet, we propose \textit{wavelet policy} to learn optimal actions based on an agent's observations from a signal processing perspective. To be more specific, multi-scale decomposition produces the representations of an input signal at different time scales, which means that wavelet transform breaks down a complex long-term sequence into different frequency components, providing a high-level overview of the \textit{overall} trend (\ie, low-frequency components) of the sequence as well as capturing \textit{detailed} changes (\ie, high-frequency components) simultaneously. By analyzing global and local characteristics of the sequence, wavelets policy maintains consistency over long horizons and also focuses on critical changes at finer time resolutions. Furthermore, by decomposing the input signals into different frequency components and isolating low-frequency components from high-frequency components which usually contain noise, wavelet policy can enhance the visibility of distinct action modes embedded in the data. In addition, wavelet policy could also contribute to high precision by providing localized detail, as the high-frequency components obtained from wavelet decomposition often represent rapid changes in the data. 

More importantly, learning actions in a coarse to fine manner by adding details gradually not only lets the policy learn actions with more accuracy but also eases the learning process. As illustrated in Figure~\ref{fig:teaser}, the sequence of images in the left panel depicts a robot arm executing a sequence of actions in a kitchen scene~\cite{gupta2020relay}. The middle and right panels show the decomposition of actions for the arm's 8$^{th}$ and 9$^{th}$ joints, respectively. Specifically, we break down each action sequence (blue) using Haar wavelet at four different scales, labeled as $f\text{-}1$ to $f\text{-}4$ (red to orange), representing the transition from finer to coarser components. We notice that the curves of both joints at the coarsest level $f\text{-}4$ reveal several distinct modes and are smoother and less fluctuating than those at the finer levels. This indicates a dominant, stable trend in the action of the joints, possibly representing a key movement or trajectory segment necessary for successful task completion. Moreover, to generate the fine-grained actions, we can start by generating the smoothest actions from $f\text{-}4$ scale and then add details gradually to the smoothest action sequence, which conceptually shares similarities with residual connection~\cite{he2016deep}.

However, another two problems arise when directly integrating wavelet analysis into wavelet policy. First, manually selecting a certain type of wavelet, \eg, Haar, Daubechies, or Morlet, is heuristic, which could be less accurate. Second, the traditional wavelet analysis is a non-learnable process that lacks flexibility and generalizability to different action prediction tasks. To resolve these two problems, we design our wavelet policy based on \textit{lifting scheme}~\cite{sweldens1998lifting}, a technique for both designing wavelets and performing discrete wavelet transform. Our wavelet policy can be learned in an end-to-end manner, inheriting the advantages from both wavelet analysis and neural networks. We conduct experimental evaluations of our wavelet policy across five different simulation environments containing different complex tasks, including but not limited to self-driving simulations~\cite{dosovitskiy2017carla}, sequential task completion in kitchen environments~\cite{gupta2020relay}, multi-robot collaboration~\cite{mandlekar2022matters}, \etc. Our contributions are summarized as follows:
\begin{enumerate}
    \item To the best of our knowledge, it is the first attempt to introduce wavelet analysis to policy learning from a signal processing perspective.
    \item We design a learnable wavelet policy network inspired by lifting scheme and provide some key design insights, along with one instantiation of this network.
    \item We evaluate our wavelet policy on five benchmarks of long-horizon tasks and achieve superior or comparable performance to the compared baselines.
\end{enumerate}
%

\section{Related Work}
\label{sec:related}

\noindent \textbf{Imitation learning.} 
Learning from expert demonstrations is an effective way to enable an agent to acquire the ability to interact with environments. OTR~\cite{luooptimal} assigns rewards to offline trajectories by using optimal transport to align unlabeled trajectories with expert demonstrations, generating an interpretable reward signal. PIRLNav~\cite{ramrakhya2023pirlnav} proposes a two-stage navigation policy learning approach that combines behavior cloning pretraining on human demonstrations with reinforcement learning fine-tuning. Diffusion policy~\cite{chi2023diffusion} leverages a conditional denoising diffusion process~\cite{ho2020denoising} to model multi-modal action distributions via a learned gradient field, and~\cite{ze20243d} extends diffusion policy to 3D cases where the denoising process is trained on 3D point cloud scenes. BESO~\cite{reussgoal} introduces a policy representation for goal-conditioned imitation learning using a score-based model~\cite{karras2022elucidating}, enabling efficient goal-specified behavior generation. SDP~\cite{wangsparse} proposes a multi-task learning approach that uses Mixture of Experts~\cite{cai2024survey} within a Transformer-based diffusion model to activate specific experts for each task. EMMA~\cite{yang2024embodied} learns to adapt a multi-modal agent in a visual world by distilling corrective feedback from a large language model. IBC~\cite{florence2022implicit} proposes implicit behavioral cloning with energy-based models~\cite{lecun2006tutorial,arbel2021generalized} for supervised policy learning, surpassing traditional explicit models.

\noindent \textbf{Learning for long-horizon tasks.} 
Learning for long-horizon tasks focuses on developing robust policies to enable effective planning and task execution over extended temporal horizons. RPL~\cite{gupta2020relay} proposes a two-phase relay policy learning algorithm that uses a hierarchical, goal-conditioned approach with a high-level policy to set subgoals and a low-level policy to act over shorter subgoal intervals. MIMICPLAY~\cite{wang2023mimicplay} first uses human play data to train a goal-conditioned latent planner that generates high-level task guidance and then utilizes limited teleoperation data to train a robot policy that follows the latent plans from the pre-trained planner. SPIRE~\cite{zhouspire} delegates complex actions to human teleoperators to collect demonstrations, which are then used to train an imitation learning policy. UVD~\cite{zhang2024universal} identifies visual subgoals in long-horizon manipulation tasks by detecting phase shifts in pre-trained visual embedding spaces to facilitate goal-based reward shaping. GSC~\cite{mishra2023generative} trains individual skill diffusion models and combines them linearly at test time to solve unseen long-horizon tasks. Hejna \etal~\cite{hejna2023improving} leverages language as auxiliary supervision to train agents with an instruction prediction loss, enhancing performance in complex, long-horizon planning tasks with limited data. Similarly, IGE-LLMs~\cite{triantafyllidis2024intrinsic} uses large language models as an intrinsic reward source to guide exploration in sparse, long-horizon manipulation tasks. DHRL~\cite{lee2022dhrl} decouples the horizons of high-level and low-level policies and adopts a graph to bridge the difference, enabling longer temporal abstraction and faster training. Behavior Transformer~\cite{shafiullah2022behavior} adapts Transformer~\cite{vaswani2017attention} with action discretization, allowing effective multi-modal continuous action prediction in long-horizon tasks.

\noindent \textbf{Wavelet for vision.} Wavelet decomposes input signals to multi-scale frequencies and has been widely used in vision tasks such as image denoising, 3D shape representation, generative modeling, \etc. MWDCNN~\cite{tian2023multi} proposes a multi-stage image denoising convolution that incorporates cascaded wavelet transform for effective noise suppression and detail recovery. WIRE~\cite{saragadam2023wire} introduces a novel activation function based on the complex Gabor wavelet, enabling highly accurate and robust implicit neural representations. AWT-Net~\cite{huang2022adaptive} proposes a multi-scale wavelet-based approach for 3D shape representation learning by hierarchically decomposing 3D shapes into frequency sub-bands using a lifting scheme~\cite{sweldens1998lifting}. WSGM~\cite{guth2022wavelet} accelerates score-based generative models~\cite{song2021score} by factorizing the data distribution into conditional probabilities of wavelet coefficients across scales, allowing efficient image synthesis. Hui \etal~\cite{hui2022neural} proposes a denoising diffusion process for 3D shape generation in wavelet domain with a compact representation using coarse and detail coefficients to model shapes. Phung \etal~\cite{phung2023wavelet} proposes to decompose and adaptively process low- and high-frequency components at both pixel and feature levels for fast image generation. WaveNeRF~\cite{xu2023wavenerf} introduces wavelet-based radian fields for high-quality and generalizable novel view synthesis by integrating wavelet frequency decomposition into multi-view stereo images. Wave-ViT~\cite{yao2022wave} integrates wavelet transforms within self-attention, enhancing self-attention outputs through inverse wavelet transforms to improve efficiency and accuracy in different vision tasks.

\section{Method}
\label{sec:method}
In this section, we first briefly review wavelet and lifting scheme in Section~\ref{subsec:lift}, and then propose our wavelet policy in Section~\ref{subsec:policy}. Next, we describe some key designs in our wavelet policy in Section~\ref{subsec:design}, and end with the loss functions for optimizing our policy in Section~\ref{subsec:loss}.

\subsection{Wavelet and Lifting Scheme}
\label{subsec:lift}
\begin{figure}[t]
\centering
\includegraphics[width=.99\linewidth]{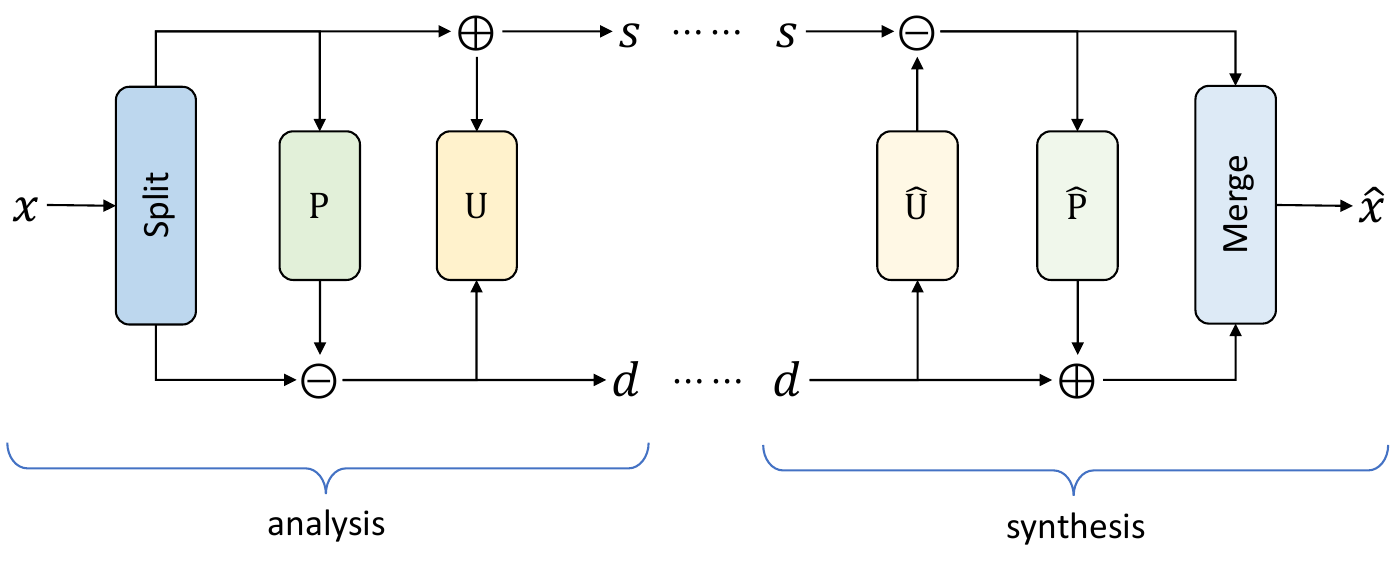}
\caption{Lifting scheme shows the analysis stage (left) where the input $x$ is split and processed by predict ($P$) and update ($U$) functions to produce detail $d$ and approximation $s$ components, and the synthesis stage (right) where inverse functions ($\hat{U},\hat{P}$) and merging operation reconstruct the original signal as $\hat{x}$.}
\label{fig:lifting}
\end{figure}
Wavelet analysis~\cite{daubechies1993ten} is a mathematical tool used to represent signals at multiple resolutions by decomposing the signals into high- and low-frequency components, which has proven essential for numerous applications in signal processing~\cite{rioul1991wavelets,mallat1999wavelet,sundararajan2016discrete}. The classical wavelet transform involves two core functions: scaling function (or low-pass filter) $\phi(x)$ and wavelet function (or high-pass filter) $\psi(x)$. The wavelet transform projects a signal $f(x)$ onto a set of scaled and translated scaling and wavelet functions:
\begin{equation}
    f(x) = \sum_kc_{j_0,k}\phi_{j_0,k}(x) + \sum_{j=j_0}\sum_kd_{j,k}\psi_{j,k}(x)\enspace,
\end{equation}
where $c_{j_0,k}$ are \textit{approximation} coefficients, $d_{j,k}$ are \textit{detail} coefficients, and $j$ and $k$ denote scale and translation parameters, respectively. This decomposition, however, relies on convolution operations with filters derived from $\phi(t)$ and $\psi(t)$, which is computationally intensive.

\begin{figure*}[t]
\centering
\includegraphics[width=.99\linewidth]{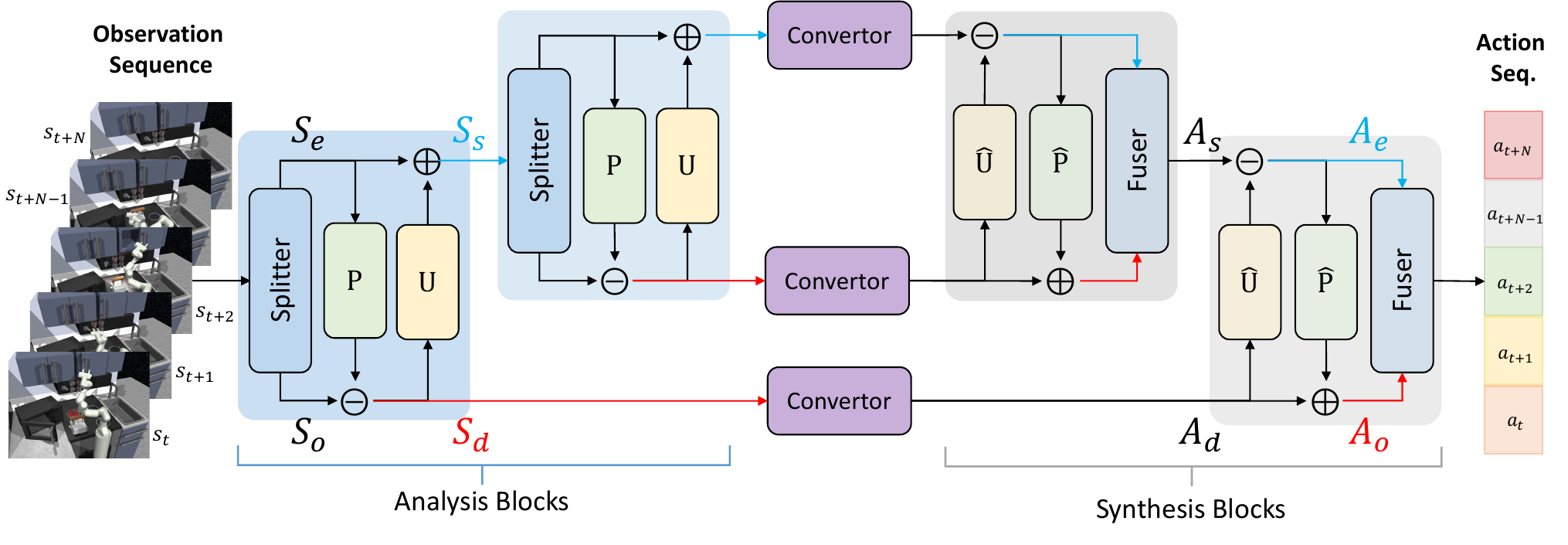}
\caption{An illustration of wavelet policy network with a multi-scale lifting scheme, where an observation sequence (left) is processed through splitting, prediction (P), and update (U) steps recursively, with converters and fusers recombining components to generate a corresponding action sequence (right). Blue lines and arrows indicate low-frequency  (coarse scale) components while red ones indicate high-frequency (fine-scale) components. Here, we only plot two scales for illustration.}
\label{fig:pipeline}
\end{figure*}

The lifting scheme~\cite{sweldens1998lifting,daubechies1998factoring} is an alternative, computationally efficient way to compute wavelet transforms. This scheme consists of three main steps, as shown in the left panel of Figure~\ref{fig:lifting}:
\begin{itemize}
    \item \textbf{Split}: Divide the signal $x[n]$ into even and odd samples, \ie, $x_\texttt{even} = x[2n]$ and $x_\texttt{odd} = x[2n+1]$.
    \item \textbf{Predict}: Use the even samples to predict the odd samples, capturing high-frequency \textit{details}:
    \begin{equation}
        d = x_\texttt{odd} - P(x_\texttt{even})\enspace.
        \label{eq:detail}
    \end{equation}
    \item \textbf{Update}: Use the predicted details to update the even samples, capturing low-frequency \textit{approximation}:
    \begin{equation}
        s = x_\texttt{even} + U(d)\enspace.
        \label{eq:approx}
    \end{equation}
\end{itemize}
Note that $P$ and $U$ are the predict and update functions, respectively. For instance, if we set $P(x) = x$ and $U(x) = \frac{x}{2}$, we can recover the Haar wavelet transform. The lifting scheme maintains the core characteristics of classical wavelets while providing improved computational efficiency. Another key advantage of lifting scheme is that we can design $P$ and $U$ as \textit{adaptive or learnable} functions, as adopted in~\cite{claypoole2003nonlinear,rodriguez2020deep,huang2022adaptive}, to increase the flexibility of the transforms. However, these previous methods only focus on the \textit{analysis} (or \textit{decomposition}) stage while ignoring the \textit{synthesis} (or \textit{reconstruction}) stage, as shown in the right panel of Figure~\ref{fig:lifting}. The synthesis stage also contains three main steps, as symmetric to the analysis stage:
\begin{itemize}
    \item \textbf{Inverse Update}: Use details to reconstruct the even samples by removing the influence of $d$ from $s$:
    \begin{equation}
        \hat{x}_\texttt{even} = s - \hat{U}(d)\enspace.
        \label{eq:even}
    \end{equation}
    \item \textbf{Inverse Predict}: Approximate the odd samples based on the even samples:
    \begin{equation}
        \hat{x}_\texttt{odd} = d + \hat{P}(\hat{x}_\texttt{even})\enspace.
        \label{eq:odd}
    \end{equation}
    \item \textbf{Merge}: Interleave the reconstructed even and odd samples to recover the original signal, \ie, $\hat{x}[2n] = \hat{x}_\texttt{even}$ and $\hat{x}[2n+1] = \hat{x}_\texttt{odd}$.
\end{itemize}
Provided that there are no numerical inaccuracies or data loss during computation, the input signal can be always perfectly reconstructed with $P = \hat{P}$ and $U = \hat{U}$.

\subsection{Wavelet Policy}
\label{subsec:policy}

We design a \textit{wavelet policy} rooted in the lifting scheme described in Section~\ref{subsec:lift} to transform an observation sequence $S= \{s_t, \cdots, s_{t+N}\}$ into a corresponding action sequence $A=\{a_t, \cdots, a_{t+N}\}$ where $N$ is the window size (or context/history size), as shown in Figure~\ref{fig:pipeline}. Given an observation sequence $S$, a \textit{splitter} decomposes the observation sequence into two streams $S_e^1$ and $S_o^1$ where the superscript $l = \{1, 2, \cdots, L\}$ denotes the level (or scale), and these streams are further processed through a learnable \textit{P-Net} and a learnable \textit{U-Net}, resulting in $S_s^1$ and $S_d^1$ streams. Distinct from the single scale scheme illustrated in Figure~\ref{fig:lifting}, we further forward $S_s^1$ into another splitter to repeat this process. Then, we adopt a series of \textit{converters} to convert $S_s^L$ and $\{S_d^l\}$ into $A_s^L$ and $\{A_d^l\}$. The converted two streams are undergone through another series of learnable \textit{$\hat{\text{U}}$-Net} and \textit{$\hat{\text{P}}$-Net}, and recombined into an action sequence by multiple $\textit{fusers}$, which are also learnable networks.

Formally, we consider each analysis block defined as:
\begin{equation}
    S_d = S_o - \mathcal{P}(S_e, \{W_i\})\enspace, S_s = S_e + \mathcal{U}(S_d, \{W_i\})\enspace,
    \label{eq:l_analysis}
\end{equation}
where the $W_i$ represents network weights to be learned. Similarly, a synthesis block is defined as:
\begin{equation}
    A_e = A_s - \hat{\mathcal{U}}(A_d), \{W_j\})\enspace, A_o = A_d + \hat{\mathcal{P}}(A_e, \{W_j\})\enspace.
    \label{eq:l_synthesis}
\end{equation}
The plus/minus operations in Eq.~\ref{eq:l_analysis} and Eq.~\ref{eq:l_synthesis} are element-wise addition and subtraction. We omit superscript for conciseness. Instead of the traditional merge operation based on positional interleaving described in Section~\ref{subsec:lift}, we employ a learnable fuser to mix the two streams $A_e$ and $A_o$ at each scale. Note that the form of $\mathcal{P}$, $\mathcal{U}$, $\hat{\mathcal{P}}$ and $\hat{\mathcal{U}}$ is flexible, provided satisfying one requirement as described below.

\subsection{Key Designs}
\label{subsec:design}

\noindent \textbf{Issues of vanilla lifting architecture.}
First, temporal causality refers to the principle that current and future states/observations (or actions) are only causally influenced by past and present states, which is important in policy learning. However, the scheme expressed in Eq.~\ref{eq:detail} to Eq.~\ref{eq:odd} does not necessarily always satisfy the causality requirement, provided no constraint on $\mathcal{P}$, $\mathcal{U}$, $\hat{\mathcal{P}}$ and $\hat{\mathcal{U}}$. Second, due to the even-odd split operation in each analysis block, for a lifting structure with $L$ scales, the input sequence at least has to be $2^L$ in length, if no padding is considered at each scale. This issue weakens the flexibility of the structure in processing input sequences with variable lengths. Third, if we adopt \textit{redundant lifting scheme}, a variant of the basic lifting scheme described in Section~\ref{subsec:lift} that modifies the traditional approach by not splitting the input sequence strictly into even and odd samples, the merge operation will be problematic, as it doubles the length of the input.

\begin{figure}[t]
\centering
\includegraphics[width=.99\linewidth]{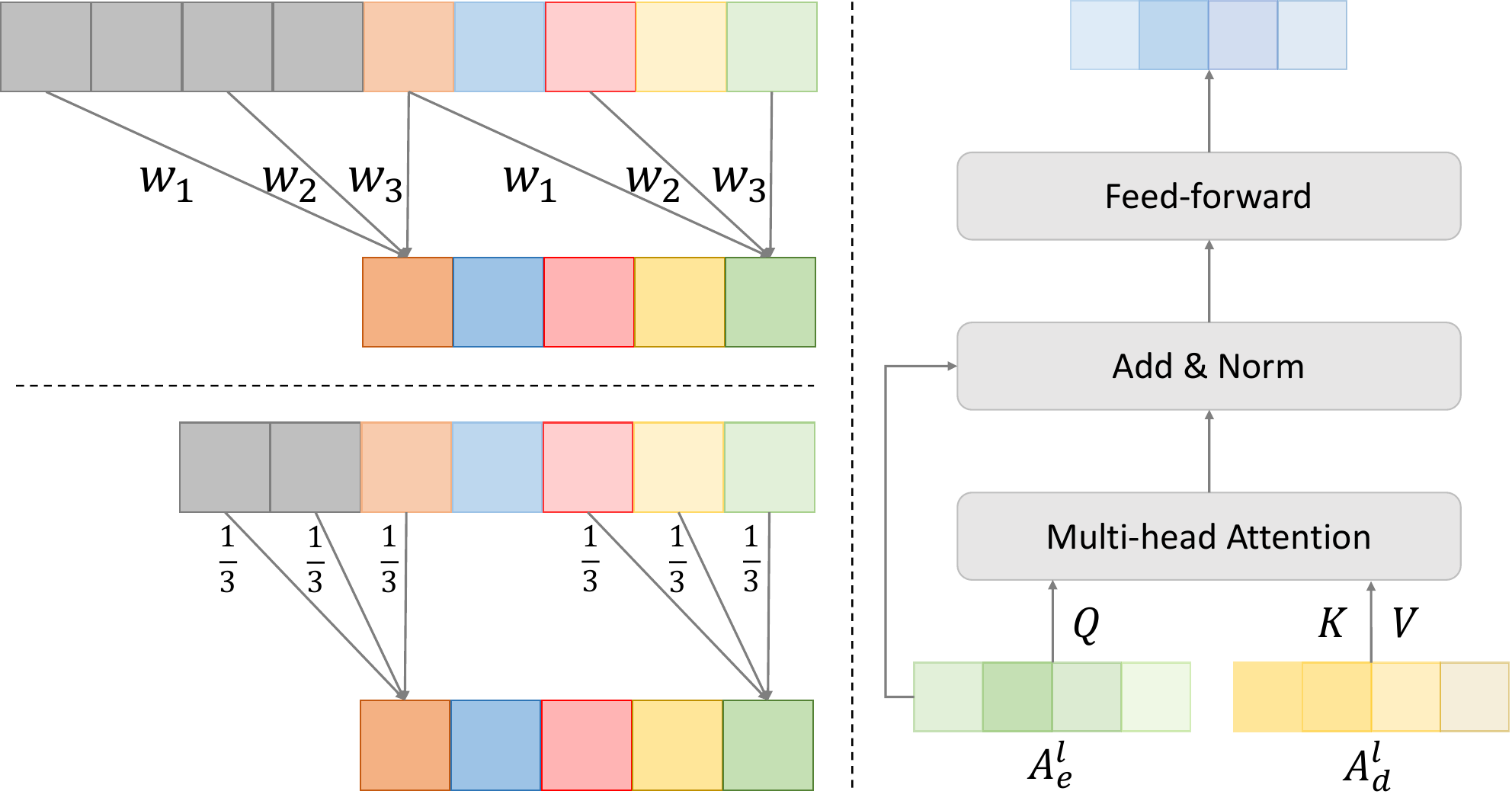}
\caption{Illustration of: causal dilated convolution (top-left) to instantiate $\mathcal{P}$, $\mathcal{U}$, $\hat{\mathcal{P}}$ and $\hat{\mathcal{U}}$; causal moving average (bottom-left); Transformer-based structure of fusers (right).}
\label{fig:design}
\end{figure}

\noindent \textbf{Solutions.}
To resolve the issues mentioned above, we first instantiate $\mathcal{P}$, $\mathcal{U}$, $\hat{\mathcal{P}}$ and $\hat{\mathcal{U}}$ as causal convolution~\cite{van2016wavenet} with dilation, which is fundamentally used in temporal signal processing. It ensures that the output at time $t$ is computed only from inputs at times $t$ and earlier, preventing the model from violating the temporal order where future inputs influence past and present outputs, as shown in the top-left panel in Figure~\ref{fig:design}. For the second issue, we adopt the redundant lifting scheme, \ie, we instantiate the splitters as Transformers with self-attention and copy the output to both ``even'' and ``odd'' streams. However, if we examine the Eq.~\ref{eq:approx} to Eq.~\ref{eq:odd}, it is worth noting that the input elements that are temporally spaced one apart from each other are processed together within $\mathcal{P}$, $\mathcal{U}$, $\hat{\mathcal{P}}$ and $\hat{\mathcal{U}}$. Therefore, we introduce dilation into the causal convolution. Another benefit is that it allows the network to integrate input information over wider time intervals without increasing the number of parameters or the computational complexity. To address the third issue, instead of applying interleaving, we adopt a Transformer with cross-attention as the fuser to recombine the two streams together, as shown in the right panel of Figure~\ref{fig:design}. We set $Q=A_s^l$ and $K=V=A_d^l$, such that it is consistent with the concept of \textit{reconstructing a signal from coarse to fine by adding high-frequency (details) to low-frequency (approximation) hierarchically}.

\noindent \textbf{Converter.}
As shown in the middle of Figure~\ref{fig:pipeline}, we introduce another (learnable) subnetwork, dubbed a \textit{converter}, for each stream. The role of the converter is to explicitly convert the streams from observation space to action space. The concrete instantiations of the converter are flexible, \eg, causal convolution, attention with causal masks, and multi-layer perceptrons, provided they satisfy the temporal causality requirement. We can also instantiate the converter as an identity transformation, assuming the conversion between the two spaces is implicit. In our experiment, we adopt causal convolution as a convertor.

%

\subsection{Loss Functions}
\label{subsec:loss}

\noindent \textbf{Task-specific terms.}
Given that we have access to the ground-truth action sequence from expert demonstrations, the task-specific loss is straightforward, such as cross-entropy loss or mean squared error loss. Here, we denoted as $\mathcal{L}_\texttt{task}$, and its concrete form is determined by the tasks.

\noindent \textbf{Wavelet-related terms.}
Different from the conventional manually designed $P$, $U$, $\hat{P}$, and $\hat{U}$ that can achieve frequency decomposition effects, these learnable networks need optimization to form a valid lifting scheme. Specifically, the approximation stream primarily captures the low-frequency components, which represent the coarse information or the general trend of the data, thus, we enforce each even/approximation stream has the same \textit{local} average as its corresponding stream in the upper scale:
\begin{equation}
    \mathcal{L}_\texttt{approx} = \sum_{i=t}^{t+N}\text{Smooth}_{L_1}(\mathcal{C}(A_s^{l}) - A_s^{l+1})\enspace,
\end{equation}
where $\mathcal{C}(\cdot)$ denotes \textit{causal moving average}, as shown in the left-bottom panel in Figure~\ref{fig:design}. The detail stream should, on the one hand, preserve high-frequency components, and on the other hand, not overpower the approximation stream. Following~\cite{huang2022adaptive}, we adopt the loss as:
\begin{equation}
    \mathcal{L}_\texttt{detail} = \sum_{i=t}^{t+N}\text{Smooth}_{L_1}(A_d^l)\enspace.
\end{equation}
The total loss for optimization is the sum of all the three above loss terms:
\begin{equation}
    \mathcal{L} = \mathcal{L}_\texttt{approx} + \alpha \mathcal{L}_\texttt{approx} + \beta \mathcal{L}_\texttt{detail}\enspace,
\end{equation}
where $\alpha$ and $\beta$ balance the weights of the last two terms.

\section{Experiments}
\label{sec:exp}

\subsection{Environments and Datasets}
\label{subsec:env}
\noindent \textbf{CARLA self-driving.} CARLA~\cite{dosovitskiy2017carla}, build on Unreal Engine~\cite{unreal2024engine}, creates a visually realistic simulated driving environment. The agent's action space is two-dimensional, allowing for acceleration/braking and left/right steering. The observation space consists of $224 \times 224$ RGB images captured from the car's perspective. In total, one hundred demonstration episodes navigate a building block in two opposite directions. Following~\cite{shafiullah2022behavior}, we use a frozen ResNet-18 encoder~\cite{he2016deep}, pretrained on ImageNet~\cite{deng2009imagenet}, as the visual encoder to encode the observations into real-value vectors.


\noindent \textbf{Franka kitchen.} To showcase the challenge of completing long-horizon tasks, we employ the Relay Kitchen Environment~\cite{gupta2020relay}, where a Franka robot~\cite{franka2024robotics} operates within a simulated kitchen setting. The kitchen contains seven interactive objects: a microwave, kettle, sliding cabinet, hinged cabinet, light switch, and two burner knobs. This environment contains a dataset of 566 demonstrations gathered from human participants using VR headsets to complete a series of four object-interaction tasks~\cite{gupta2020relay,shafiullah2022behavior} with these objects.

\noindent \textbf{Push-T.} Push-T is derived from IBC~\cite{florence2022implicit}, aiming at pushing a T-shaped block toward a designated target using a circular end-effector. The task introduces variability through randomized starting positions for both the T-block and the end-effector. Successfully completing the task demands precise manipulation of the block by leveraging intricate, contact-heavy object dynamics via point contacts. Evaluations were conducted using 9 keypoints in a 2D space, extracted from the ground-truth pose of the T-block, alongside proprioceptive feedback for the end-effector’s position.

\noindent \textbf{Transport.} We select a difficult multi-robot collaboration task, \ie, Transport, from Robomimic~\cite{mandlekar2022matters}, a robotic manipulation simulated environment for imitation learning and offline reinforcement learning evaluation. One variant of this task contains teleoperated demonstrations from a skilled operator (PH), while another one includes mixed demonstrations performed by both skilled and less skilled operators (MH). We present performance outcomes based on the state information inputs.

\noindent \textbf{D3IL.} The D3IL simulation environment~\cite{jia2024towards} is designed to evaluate policy learning algorithms on tasks that require diverse behaviors, multi-object manipulation, and closed-loop sensory feedback. The environment consists of five tasks: Avoiding, Aligning, Pushing, Sorting, and Stacking, each with distinct challenges such as obstacle avoidance, object alignment, multi-object sorting, and dexterous stacking. These tasks are designed to incorporate multiple valid policies, ensuring that models must learn multi-modal behavior rather than a single deterministic policy. 

\subsection{Implementation}
\label{subsec:impl} We implement two variants of our wavelet policy: one integrated into Behavior Transfomer (BeT)~\cite{shafiullah2022behavior}, and one integrated into Diffusion Policy with Transformer~\cite{chi2023diffusion}. For the former one, we replace MinGPT~\cite{brown2020language} with our wavelet network while keeping all other components unchanged, ensuring a fair comparison. For the latter one, we replace the series of Transformer-based~\cite{vaswani2017attention} decoders with our wavelet network while still using the Transformer-based encoders to encode the input to high dimensional features. We also keep the optimizer, scheduler, learning rate, batch size as the same values used in the original BeT and Diffusion Policy papers. We set the hyperparameters $\alpha$ and $\beta$ to 0.1. All experiments are conducted with three different random seeds, and the average values and standard derivations of each metric are reported.

\subsection{Evaluation Results}
\label{subsec:result}

\noindent \textbf{CARLA self-driving} and \textbf{Franka kitchen.} Using the same metrics as~\cite{shafiullah2022behavior}, in the CARLA task, success is evaluated by the probability that the car reaches its target destination, and evaluations are conducted over 100 rollouts. 
For the Kitchen environment, performance is measured by the probability that the model completes up to $N$ tasks (from T1 to T5), \ie, interacts with up to $N$ objects successfully, within a limit of 280 timesteps. In total, 1,000 rollouts are evaluated for the Kitchen environment.

%
\begin{table*}[!h]
\small
\centering
\begin{tabular}{l|c|ccccc}
\toprule
 & \multicolumn{1}{c|}{\textbf{CARLA}} & \multicolumn{5}{c}{\textbf{Kitchen}} \\
 & Success & T1 & T2 & T3 & T4 & T5 \\
\midrule
BeT~\cite{shafiullah2022behavior} & 0.832\footnotesize{$\pm$0.167}  & 0.948\footnotesize{$\pm$0.034} & 0.773\footnotesize{$\pm$0.065} & 0.562\footnotesize{$\pm$0.095} & 0.275\footnotesize{$\pm$0.066} & 0.027\footnotesize{$\pm$0.023} \\
VQ-BeT~\cite{lee2024behavior} & 0.839\footnotesize{$\pm$0.125}  & 0.950\footnotesize{$\pm$0.021} & \textbf{0.775}\footnotesize{$\pm$0.046} & 0.559\footnotesize{$\pm$0.105} & 0.306\footnotesize{$\pm$0.057} & 0.029\footnotesize{$\pm$0.020} \\
\midrule
\textbf{Ours} & \textbf{0.847}\footnotesize{$\pm$0.090} & \textbf{0.953}\footnotesize{$\pm$0.020} & \textbf{0.775}\footnotesize{$\pm$0.057} & \textbf{0.563}\footnotesize{$\pm$0.063} & \textbf{0.339}\footnotesize{$\pm$0.071} & \textbf{0.041}\footnotesize{$\pm$0.027} \\
\bottomrule
\end{tabular}
\caption{Quantitative results on CARLA and Franka kitchen tasks.}
\label{tab:two_in_one}
\end{table*}

The quantitative results are shown in Table~\ref{tab:two_in_one}. Our approach surpasses or closely matches the baselines BeT and VQ-BeT in all tasks in both CARLA and the Franka kitchen environment. In CARLA, we achieved a higher success rate (0.847 $\pm$ 0.090) compared to 0.832 $\pm$ 0.167 for BeT and 0.839 $\pm$ 0.125 for VQ-BeT, indicating a more reliable ability to reach the target destination across 100 rollouts. In the Kitchen tasks, performance is very similar for T1 to T3, but our method shows a noticeable improvement for T4 and T5, which suggests that our approach excels at handling more complex or long-horizon interactions. The reduced standard deviations in several cases (\eg, CARLA and T1 in Kitchen) also indicate stronger consistency of our results.

\begin{figure*}[!t]
\centering
\includegraphics[width=.88\linewidth]{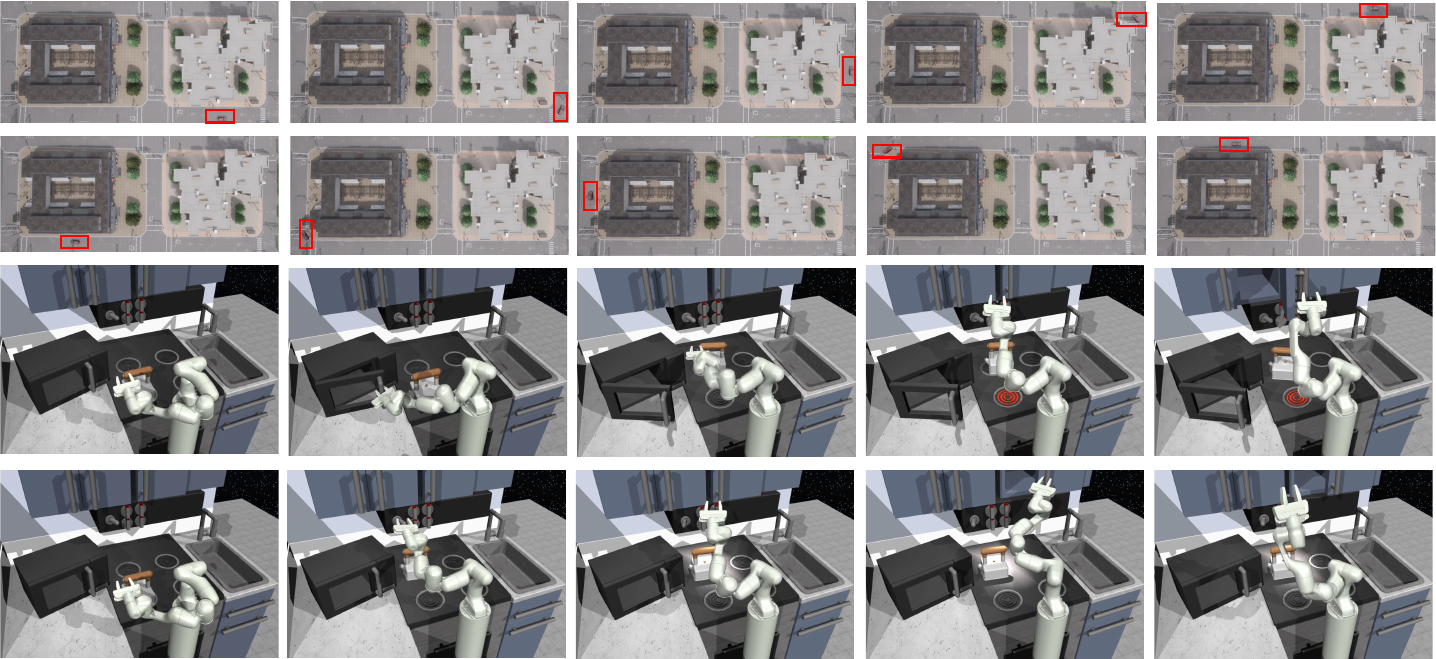}
\caption{Visualization of agent trajectories in two benchmarks. The car (highlighted by a red box) in CARLA drives right (1$^{th}$ row) and left (2$^{nd}$ row). A Franka Arm accomplishes the task of Microwave - Kettle - Bottom knob - Hinge (5$^{th}$ row) and another task of Kettle - Light switch - Slide (6$^{th}$ row). Note that the last figure in the last row shows that the Arm tries to open the hinge, but fails.}
\label{fig:vis}
\end{figure*}

Qualitative visualizations on the two tasks are shown in Figure~\ref{fig:vis}. In the CARLA benchmark (1$^{st}$ and 2$^{nd}$ rows), the car, indicated by a red bounding box, successfully navigates through a simulated environment by driving to the right in the 1$^{st}$ row and driving to the left in the 2$^{nd}$ row. 
The Kitchen task (5$^{th}$ and 6$^{th}$ rows), features a Franka robotic arm performing complex interactions with kitchen appliances. The arm successfully completes tasks like manipulating the microwave, kettle, bottom knob, and hinge, showing the model’s competence in handling multi-step, object-specific interactions. 

\noindent\textbf{Multi-modal analysis.} Following the protocols proposed in~\cite{shafiullah2022behavior}, for the CARLA task, we assess the likelihood of the car turning left versus right at intersections, excluding any out-of-distribution rollouts. 
For the Franka Kitchen task, we measure the empirical entropy of task sequences generated by the model, treating these sequences as strings. The results are shown in Table~\ref{tab:multi_modal} in which the values closest to those observed in the demonstrations are highlighted.
As shown in Table~\ref{tab:multi_modal}, our approach more closely recovers the multi-modal behavior distribution exhibited in the demonstrations for both CARLA and the Kitchen setting. In CARLA, the relative frequency of taking left versus right turns is closer to the 50–50 split observed in the demonstration data. In the Kitchen environment, our model’s generated task sequences also achieve a higher entropy than BeT and VQ-BeT, bringing us slightly nearer to the 2.96 entropy level from the demonstrations. These results suggest that our approach produces behavior distributions better aligned to the diverse, multi-modal nature of the training data.

\begin{table}[!h]
\small
\centering
\begin{tabular}{l|cc|c}
\toprule
 & \multicolumn{2}{c|}{\textbf{CARLA}} & \multicolumn{1}{c}{\textbf{Kitchen}} \\
 & Left & Right & Entropy \\
\midrule
BeT~\cite{shafiullah2022behavior} & 0.293\footnotesize{$\pm$0.153} & 0.699\footnotesize{$\pm$0.154} & 2.506\footnotesize{$\pm$0.078} \\
VQ-BeT~\cite{lee2024behavior} & 0.315\footnotesize{$\pm$0.146} & 0.674\footnotesize{$\pm$0.145} & 2.508\footnotesize{$\pm$0.075} \\
Demonstrations & 0.50 & 0.50 & 2.96 \\
\midrule
\textbf{Ours} & \textbf{0.337}\footnotesize{$\pm$0.126} & \textbf{0.662}\footnotesize{$\pm$0.161} & \textbf{2.511}\footnotesize{$\pm$0.069}  \\
\bottomrule
\end{tabular}
\caption{Statistics of multi-modal behavior modes learned from different methods on CARLA and Franka kitchen tasks.}
\label{tab:multi_modal}
\end{table}

\noindent \textbf{Push-T} and \textbf{Transport.} 
As reported in Table~\ref{tab:pusht_robo}, our DP-Wavelet consistently outperforms DP-Transformer, with particularly notable gains in Transport-mh, suggesting it is more robust to variability in demonstration quality. The results imply that wavelet-based diffusion policies can effectively handle both uniform and mixed-skill data, leading to stronger overall performance. It also verifies that our wavelet policy can more effectively capture and utilize both local and global features relevant to pushing, positioning and collaboration tasks.


%
\begin{table}[!h]
\small
\setlength{\tabcolsep}{1.5pt}
\centering
\begin{tabular}{l|c|c|c}
\toprule
 & \textbf{Push-T} & \textbf{Transport-ph} & \textbf{Transport-mh} \\
\midrule
DP-Transformer~\cite{chi2023diffusion} & 0.942\footnotesize{$\pm$0.014} & 0.820\footnotesize{$\pm$0.055} & 0.440\footnotesize{$\pm$0.068} \\
\midrule
\textbf{DP-Wavelet (Ours)} & \textbf{0.958}\footnotesize{$\pm$0.016} & \textbf{0.835}\footnotesize{$\pm$-0.062} & \textbf{0.497}\footnotesize{$\pm$0.085}  \\
\bottomrule
\end{tabular}
\caption{Quantitative results of success rate on Push-T and Transport tasks. `DP' denotes Diffusion Policy.}
\label{tab:pusht_robo}
\end{table}

\noindent \textbf{D3IL.} As summarized in Table~\ref{tab:d3il}, our method offers consistent gains in success rate across all six tasks when compared to BeT, while maintaining or slightly increasing the policy entropy in most cases. In the Avoiding, Aligning, and Pushing tasks, our approach achieves notably higher success rates (\eg, 0.567$\pm$0.040 \textit{vs.} 0.463$\pm$0.163 for Avoiding) alongside entropy values close to those of BeT, indicating more reliable task execution without collapsing into a single, repetitive strategy. In Sorting‑2, both methods display very low entropy, reflecting that the task can be solved effectively by repeating the same or similar actions. For Pushing tasks, our method again exhibits also higher noticeable success rates and marginally higher entropy, suggesting it can better handle the multi-step nature of stacking while still exploring diverse strategies. These findings imply that our wavelet policy attains higher overall performance while preserving behavioral diversity, consistent with the design goal of matching the distribution of behaviors in the data collection process.

\begin{table*}[!h]
\small
\centering
\begin{tabular}{l|cc|cc|cc}
\toprule
              & \multicolumn{2}{c|}{\textbf{Avoiding}}  & \multicolumn{2}{c|}{\textbf{Aligning}}   & \multicolumn{2}{c}{\textbf{Pushing}}    \\ \midrule
              & Success Rate                 & Entropy               & Success Rate                  & Entropy                & Success Rate                  & Entropy                \\ \midrule
BeT~\cite{shafiullah2022behavior}  & 0.463\footnotesize{$\pm$0.163}    & 0.840\footnotesize{$\pm$0.059}   & 0.575\footnotesize{$\pm$0.050}    & 0.458\footnotesize{$\pm$0.095}    & 0.578\footnotesize{$\pm$0.064}    & 0.774\footnotesize{$\pm$0.043}    \\
IBC~\cite{florence2022implicit}  & 0.523\footnotesize{$\pm$0.040}    & 0.769\footnotesize{$\pm$0.036}   & 0.633\footnotesize{$\pm$0.058}    & 0.283\footnotesize{$\pm$0.101}    & 0.535\footnotesize{$\pm$0.012}    & 0.784\footnotesize{$\pm$0.036}    \\
\textbf{Ours} & \textbf{0.567}\footnotesize{$\pm$0.040}    & \textbf{0.847}\footnotesize{$\pm$0.009}   & \textbf{0.710}\footnotesize{$\pm$0.029}    & \textbf{0.541}\footnotesize{$\pm$0.041}    & \textbf{0.793}\footnotesize{$\pm$0.019}    & \textbf{0.818}\footnotesize{$\pm$0.045}    \\ \midrule\midrule
              & \multicolumn{2}{c|}{\textbf{Sorting-2}} & \multicolumn{2}{c|}{\textbf{Stacking-1}} & \multicolumn{2}{c}{\textbf{Stacking-2}} \\ \midrule
              & Success Rate                  & Entropy               & Success Rate                  & Entropy                & Success Rate                  & Entropy                \\ \midrule
BeT~\cite{shafiullah2022behavior}  & 0.693\footnotesize{$\pm$0.019}    & 0.001\footnotesize{$\pm$0.000}   & 0.188\footnotesize{$\pm$0.096}    & \textbf{0.293}\footnotesize{$\pm$0.061}    & 0.031\footnotesize{$\pm$0.010}    & 0.078\footnotesize{$\pm$0.005}    \\
IBC~\cite{florence2022implicit}  & 0.456\footnotesize{$\pm$0.030}    & \textbf{0.332}\footnotesize{$\pm$0.048}   & 0.192\footnotesize{$\pm$0.103}    & 0.285\footnotesize{$\pm$0.053}    & 0.025\footnotesize{$\pm$0.054}    & 0.076\footnotesize{$\pm$0.129}    \\
\textbf{Ours} & \textbf{0.723}\footnotesize{$\pm$0.101}    & 0.001\footnotesize{$\pm$0.000}   & \textbf{0.197}\footnotesize{$\pm$0.089}    & 0.270\footnotesize{$\pm$0.077}    & \textbf{0.041}\footnotesize{$\pm$0.027}    & \textbf{0.085}\footnotesize{$\pm$0.082}    \\ \bottomrule
\end{tabular}
\caption{Quantitative results of six tasks in D3IL simulation environment. Entropy is used to qualify the diversity of behaviors learned by the methods. The higher value, \ie, approaching 1.0, indicates more diverse modes of behavior.}
\label{tab:d3il}
\end{table*}
%

\section{Ablation Study}
\label{sec:abl}



\noindent \textbf{Effect of causal convolution.} As discussed above in Section~\ref{subsec:design}, causal dilated convolution is a key design factor. In Table~\ref{tab:conv}, we replace causal convolution with conventional non-causal convolution in all $\mathcal{P}$, $\mathcal{U}$, $\hat{\mathcal{P}}$ and $\hat{\mathcal{U}}$ and notice a substantial drop in performance across all Kitchen tasks. This underscores the importance of preserving causal order in policy learning to properly handle sequential interactions during manipulation.

\begin{table}[h]
\footnotesize
\centering
\setlength{\tabcolsep}{1.5pt}
\begin{tabular}{l|ccccc}
\toprule
 & \multicolumn{5}{c}{\textbf{Kitchen}} \\
 & T1 & T2 & T3 & T4 & T5 \\
\midrule
Conv & 0.494\scriptsize{$\pm$0.099} & 0.259\scriptsize{$\pm$0.091} & 0.112\scriptsize{$\pm$0.057} & 0.033\scriptsize{$\pm$0.022} & 0.002\scriptsize{$\pm$0.004} \\
\midrule
\textbf{Causal} & \textbf{0.953}\scriptsize{$\pm$0.020} & \textbf{0.775}\scriptsize{$\pm$0.057} & \textbf{0.563}\scriptsize{$\pm$0.063} & \textbf{0.339}\scriptsize{$\pm$0.071} & \textbf{0.041}\scriptsize{$\pm$0.027} \\
\bottomrule
\end{tabular}
\caption{Comparison with conventional non-causal convolution.}
\label{tab:conv}
\end{table}
%

%

\noindent \textbf{Effect of learnable wavelet.} We replace all the learnable networks of $\mathcal{P}$, $\mathcal{U}$, $\hat{\mathcal{P}}$ and $\hat{\mathcal{U}}$ shown in Figure~\ref{fig:pipeline} with two types of traditional, non-learnable wavelet, \ie, Haar and Daubechies 2 (DB2) wavelets, and the results are shown in Table~\ref{tab:haar_db2}. Note that we still keep Splitters and Convertors as learnable. The results suggest that using a flexible, data-driven wavelet representation enables the model to adaptively capture critical task features, whereas fixed wavelets may not align as effectively with the nuances of the underlying state transitions. 

%
\begin{table}[h]
\footnotesize
\centering
\setlength{\tabcolsep}{1.5pt}
\begin{tabular}{l|ccccc}
\toprule
 & \multicolumn{5}{c}{\textbf{Kitchen}} \\
 & T1 & T2 & T3 & T4 & T5 \\
\midrule
Haar & 0.884\scriptsize{$\pm$0.055} & 0.668\scriptsize{$\pm$0.217} & 0.535\scriptsize{$\pm$0.096} & 0.265\scriptsize{$\pm$0.078} & 0.029\scriptsize{$\pm$0.022} \\
DB2 & 0.831\scriptsize{$\pm$0.093} & 0.666\scriptsize{$\pm$0.114} & 0.473\scriptsize{$\pm$0.110} & 0.225\scriptsize{$\pm$0.079} & 0.029\scriptsize{$\pm$0.036} \\
\midrule
\textbf{Ours} & \textbf{0.953}\scriptsize{$\pm$0.020} & \textbf{0.775}\scriptsize{$\pm$0.057} & \textbf{0.563}\scriptsize{$\pm$0.063} & \textbf{0.339}\scriptsize{$\pm$0.071} & \textbf{0.041}\scriptsize{$\pm$0.027} \\
\bottomrule
\end{tabular}
\caption{Comparison with non-learnable Haar, DB2 wavelets.}
\label{tab:haar_db2}
\end{table}
%

\section{Conclusion}
\label{sec:con}
We introduce a novel wavelet policy learning framework designed to enhance the precision and reliability of robotic tasks by utilizing multi-scale wavelet transformations. This method leverages lifting scheme to decompose complex, long-term sequences of actions and observations, allowing for a nuanced understanding of behaviors. 
We evaluate method on five simulation environments across different tasks, and the experimental results are promising and demonstrate its potential
to broader application senarios.

{
    \small
    \bibliographystyle{ieeenat_fullname}
    \bibliography{main}
}

\end{document}